Original Paper

# Improved Digital Therapy for Developmental Pediatrics Using Domain-Specific Artificial Intelligence: Machine Learning Study

Peter Washington, BA, MS; Haik Kalantarian, PhD; John Kent, BA, MA; Arman Husic, BS; Aaron Kline, BS; Emilie Leblanc, MS; Cathy Hou; Onur Cezmi Mutlu, BS; Kaitlyn Dunlap, MS; Yordan Penev, MS; Maya Varma, BS; Nate Tyler Stockham, MS; Brianna Chrisman, MS; Kelley Paskov, MS; Min Woo Sun, BS; Jae-Yoon Jung, PhD; Catalin Voss, MS; Nick Haber, PhD; Dennis Paul Wall, PhD

Departments of Pediatrics (Systems Medicine) and Biomedical Data Science, Stanford University, Stanford, CA, United States

**Corresponding Author:**
Peter Washington, BA, MS
Departments of Pediatrics (Systems Medicine) and Biomedical Data Science
Stanford University
Stanford, CA
United States
Phone: 1 5126800926
Email: peterwashington@stanford.edu

## Abstract

**Background:** Automated emotion classification could aid those who struggle to recognize emotions, including children with developmental behavioral conditions such as autism. However, most computer vision emotion recognition models are trained on adult emotion and therefore underperform when applied to child faces.

**Objective:** We designed a strategy to gamify the collection and labeling of child emotion–enriched images to boost the performance of automatic child emotion recognition models to a level closer to what will be needed for digital health care approaches.

**Methods:** We leveraged our prototype therapeutic smartphone game, GuessWhat, which was designed in large part for children with developmental and behavioral conditions, to gamify the secure collection of video data of children expressing a variety of emotions prompted by the game. Independently, we created a secure web interface to gamify the human labeling effort, called HollywoodSquares, tailored for use by any qualified labeler. We gathered and labeled 2155 videos, 39,968 emotion frames, and 106,001 labels on all images. With this drastically expanded pediatric emotion–centric database (>30 times larger than existing public pediatric emotion data sets), we trained a convolutional neural network (CNN) computer vision classifier of happy, sad, surprised, fearful, angry, disgust, and neutral expressions evoked by children.

**Results:** The classifier achieved a 66.9% balanced accuracy and 67.4% F1-score on the entirety of the Child Affective Facial Expression (CAFE) as well as a 79.1% balanced accuracy and 78% F1-score on CAFE Subset A, a subset containing at least 60% human agreement on emotions labels. This performance is at least 10% higher than all previously developed classifiers evaluated against CAFE, the best of which reached a 56% balanced accuracy even when combining "anger" and "disgust" into a single class.

**Conclusions:** This work validates that mobile games designed for pediatric therapies can generate high volumes of domain-relevant data sets to train state-of-the-art classifiers to perform tasks helpful to precision health efforts.

## Introduction

Automated emotion classification can serve in pediatric care solutions, particularly to aid those who struggle to recognize emotion, such as children with autism who have trouble with emotion evocation and recognizing emotions displayed by others [1-3]. In prior work, computer vision models for emotion recognition [4-6] used in digital therapeutics have shown significant treatment effects in children with autism [7-17]. The increasing use of signals from sensors on mobile devices, such as the selfie camera, opens many possibilities for real-time analysis of image data for continuous phenotyping and repeated diagnoses in home settings [18-33]. However, facial emotion classifiers and the underlying data sets on which they are trained have been tailored to neurotypical adults, as demonstrated by repeatedly low performance on image data sets of pediatric emotion expressions [34-39].

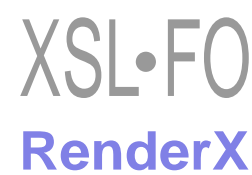

The Child Affective Facial Expression (CAFE) data set is currently the most popular facial expression data set pertaining to children. Prior machine learning efforts that do not include CAFE images in the training set have reached 56% accuracy on CAFE [36,37,39], even after combining facial expressions (eg, "anger" and "disgust") into a single class, thus limiting granularity. We do not discuss prior publications that report higher accuracy using subsets of the CAFE data set in the training and testing sets. This overall lack of performance in prior work highlights the need for developing facial emotion classifiers that work for children. With a lack of labeled data being the fundamental bottleneck to achieving clinical-grade performance, low-cost and speedy data generation and labeling techniques are pertinent.

As a first step toward the creation of a large-scale data set of child emotions, we have previously designed GuessWhat, a dual-purpose smartphone app that serves as a therapeutic for children with autism while simultaneously collecting highly structured image data enriched for emoting in children. GuessWhat was designed for children aged 2 and above to encourage prosocial interaction with a gameplay partner (eg, mom or dad), focusing the camera on the child while presenting engaging but challenging prompts for the child to try to act out [40-43]. We have previously tested GuessWhat's potential to increase socialization in children with autism as well as its potential to collect structured videos of children emoting facial expressions [44]. In addition to collecting videos enriched with emotions, GuessWhat gameplay generates user-derived labels of emotion by leveraging the charades-style gameplay structure of the therapy.

Here, we document the full pipeline for training a classifier using emotion-enriched video streams coming from GuessWhat gameplay, resulting in a state-of-the-art pediatric facial emotion classifier that outperforms all prior classifiers when evaluated on CAFE. We first recruited parents and children from around the world to play GuessWhat and share videos recorded by the smartphone app during gameplay. We next extracted frames from the videos, automatically discarding some frames through quality control algorithms, and uploaded the frames on a custom behavioral annotation labeling platform named HollywoodSquares. We prioritized the high entropy frames and shared them with a group of 9 human annotators who annotated emotions in the frames. In total, we have collected 39,968 unique labeled frames of emotions that appear in the CAFE data set. Using the resulting frames and labels, we trained a facial emotion classifier that can distinguish happy, sad, surprised, fearful, angry, disgust, and neutral expressions in naturalistic images, achieving state-of-the-art performance on CAFE and outperforming existing classifiers by over 10%. This work demonstrates that therapeutic games, while primarily providing a behavioral intervention, can simultaneously generate sufficient data for training state-of-the-art domain-specific computer vision classifiers.

## Methods

### Data Collection

The primary methodological contribution of this work is a general-purpose paradigm and pipeline (Figure 1) consisting of (1) passive collection of prelabeled structured videos from therapeutic interventions, (2) active learning to rank the collected frames leveraging the user-derived labels generated during gameplay, (3) human annotation of the frames in the order produced in the previous step, and (4) training a classifier while artificially augmenting the training set. We describe our instantiation of this general paradigm in the following sections.

**Figure 1.** Pipeline of the model training process. Structured videos enriched with child emotion evocation are collected from a mobile autism therapeutic deployed in the wild. The frames are ranked for their contribution to the target classifier by a maximum entropy active learning algorithm and receive human labels on a rating platform named HollywoodSquares. The frames are corresponding labels that are transferred onto a ResNet-152 neural network pretrained on the ImageNet data set.

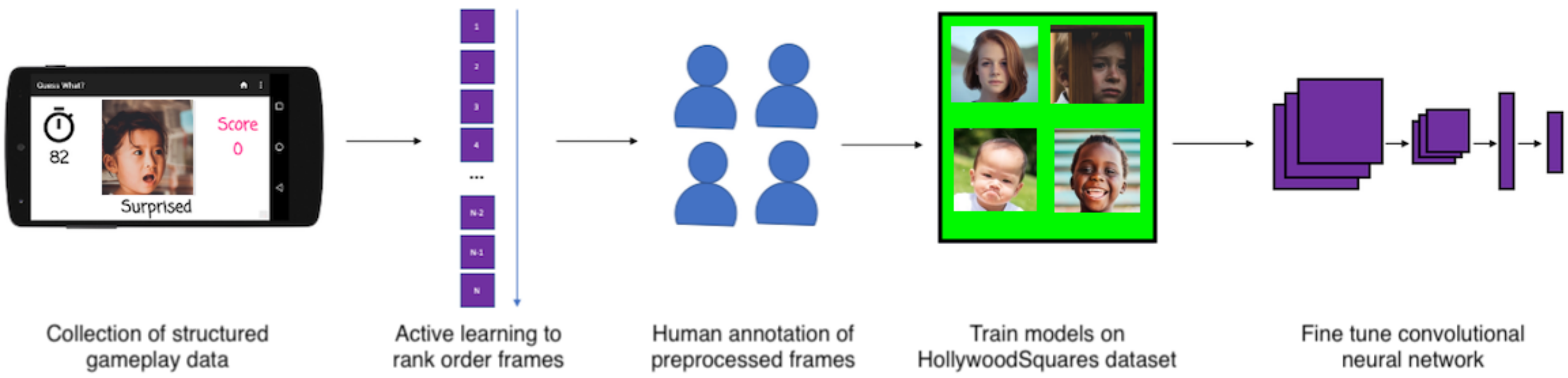

### Ethical Considerations

All study procedures, including data collection, were approved by the Stanford University Institutional Review Board (IRB number 39562) and the Stanford University Privacy Office. In addition, informed consent was obtained from all participants, all of whom had the opportunity to participate in the study without sharing videos.

## Recruitment

To recruit child video subjects, we ran a marketing campaign to gather rich and diverse video inputs of children playing GuessWhat while evoking a range of emotions. We posted advertisements on social media (Facebook, Instagram, and Twitter) and contacted prior study participants for other digital smartphone therapeutics developed by the lab [13-15]. All recruitment and study procedures were approved by the Stanford University IRB.

## User Interfaces

### GuessWhat Smartphone Therapeutic

GuessWhat is a mobile autism therapy implemented on iOS and Android, which has been previously documented as a useful tool for the collection of structured video streams of children behaving in constrained manners [40-44], including evocation of targeted emotions. GuessWhat features a charades game where the parents place the phone on their forehead facing the child, while the child acts out the emotion prompt displayed on the screen. The front-facing camera on the phone records a video of the child in addition to corresponding prompt metadata. All sessions last for 90 seconds. Upon approval by the parent, each session video is uploaded to a Simple Storage Service (S3) bucket on Amazon Web Services (AWS). The app has resulted in 2155 videos shared by 456 unique children. Parents are asked to sign an electronic consent and assent form prior to playing GuessWhat. After each gameplay session, parents can (1) delete the videos, (2) share the videos with the research team only, or (3) share the videos publicly.

## Emotions Considered

We sought labels for Paul Ekman's list of six universal emotions: anger, disgust, fear, happiness, sadness, and surprise [45-48]. Ekman originally included contempt in the list of emotions but has since revised the list of universal emotions. Because CAFE does not include labels of contempt, we did not train our classifier to predict contempt. We added a seventh category named neutral, indicating the absence of an expressed emotion. Our aim was to train a 7-way emotion classifier distinguishing among Ekman's 6 universal emotions plus neutral.

### HollywoodSquares Frame Labeling

We developed a frame-labeling website named HollywoodSquares. The website provides human labelers with an interface to speedily annotate a sequential grid of frames (Figure 2) that were collected during the GuessWhat gameplay. To enable rapid annotation, HollywoodSquares enables users to label frames by pressing hot keys, where each key corresponds to a particular emotion label. To provide a label, users can hover their mouse over a frame and press the hot key corresponding to the emotion they want to label. As more frames are collected by GuessWhat, they continue to appear on the interface. Because the HollywoodSquares system displays over 20 images on the screen at once, it encourages rapid annotation and enables simultaneous engagement by many independent labelers. This permits rapid convergence of a majority rules consensus on image labels.

We ran a labeling contest with 9 undergraduate and high school annotators, where we challenged each annotator to produce labels that would result in the highest performing classifier on the CAFE data set. Raters were aged between 15 and 24 years and were from the Bay Area, Northeastern United States, and Texas. The raters included 2 males and 7 females. For the frames produced by each individual annotator, we trained a ResNet-152 model (see Model Training). We updated annotators about the number of frames they labeled each week and the performance of the classifier trained with their individual labels. We awarded a cash prize to the annotator with the highest performance at the end of the 9-week labeling period.

**Figure 2.** HollywoodSquares rating interface. Annotators use keyboard shortcuts and the mouse to speedily annotate a sequence of frames acquired during GuessWhat gameplay.

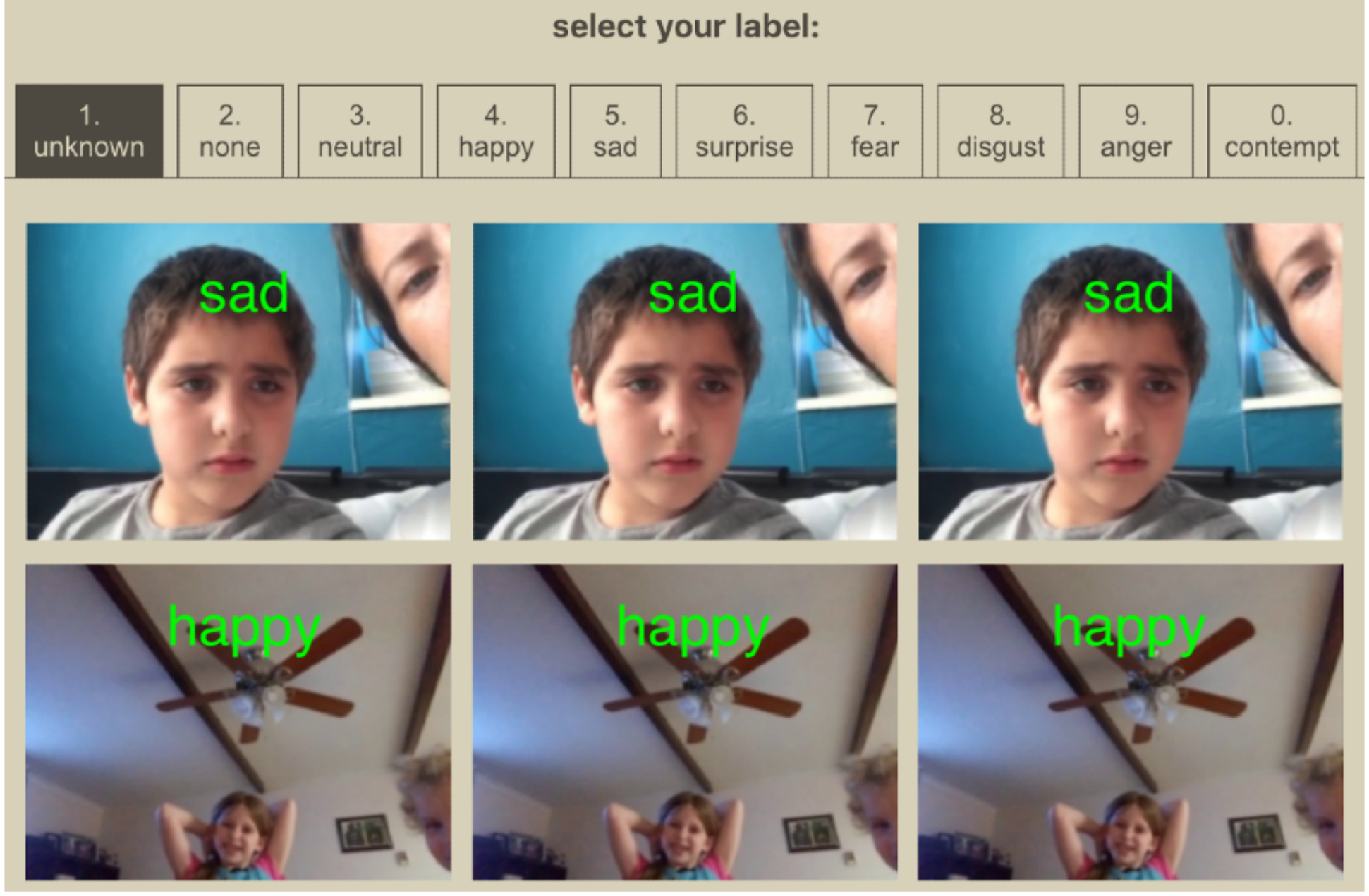

HollywoodSquares was also used for a testing phase, during which iterations of the frame-labeling practices were made between the research and annotation teams. All the labeled frames acquired during this testing phase were discarded for final classifier training.

All annotators were registered as research team members through completion of the Health Insurance Portability and Accountability Act of 1996 and Collaborative Institutional Training Initiative training protocols in addition to encrypting their laptop with Stanford Whole Disk Encryption. This provided annotators with read-only access to all the videos and derived frames from GuessWhat gameplay that were shared with the research team.

The final labels were chosen by the following process. If all annotators agreed unanimously about the final frame label, then this label was assigned as the final frame label. If disagreements existed between raters, then the emotion gameplay prompt associated with that frame (the "automatic label") was assigned as the final label for that frame, as long as at least 1 of the human annotators agreed with the automatic label. If disagreements existed between raters but the automatic label did not match any human annotations, then the frame was not included in the final training data set.

### Machine Learning

#### *Model Training*

We leveraged an existing CNN architecture, ResNet-152 [49], with pretrained weights from ImageNet [50]. We used categorical cross entropy loss and Adam optimization with a learning rate of $3 \times 10^{-4}$, with $\beta_1$ set to .99 and $\beta_2$ set to .999. We retrained every layer of the network until the training accuracy converged. The model converged when it did not improve against a validation data set for 20 consecutive epochs. We applied the following data augmentation strategies in conjunction and at random for each training image and each batch of training: rotation of frames between –15 and 15 degrees, zooming by a factor between 0.85 and 1.15, shifting images in every direction by up to 1/10th of the width and height, changing brightness by a factor between 80% and 120%, and potential horizontal flipping.

The CNN was trained in parallel on 16 graphics processing unit (GPU) cores with a p2.16xlarge Elastic Cloud Compute instance on AWS using the Keras library in Python with a Tensorflow 2 backend. With full GPU usage, the training time was 35 minutes and 41 seconds per epoch for a batch size of 1643, translating to US $14.4 per hour.

We trained 2 versions of the model, with 1 exclusively using non-GuessWhat public data set frames from (1) the Japanese Female Facial Expression (JAFFE) [51], (2) a random subset of 30,000 AffectNet [52] images (a subset was acquired to avoid an out of memory error), and (3) the Extended Cohn-Kanade (CK+) data set [53]; the other model was trained with these public data set frames plus all 39,968 labeled and relevant GuessWhat frames.

#### *Model Evaluation*

We evaluated our models against the entirety of the CAFE data set [54], a set of front-facing images of racially and ethnically diverse children aged 2 to 8 years expressing happy, sad, surprised, fear, angry, fearful, and neutral emotions. CAFE is currently the largest data set of facial expressions from children and has become a standard benchmark for this field.

Although existing studies have evaluated models exclusively against the entirety of the CAFE data set [34-39], we additionally evaluated them on Subset A and Subset B of CAFE, as defined by the authors of the data set. Subset A contains images that were identified with an accuracy of 60% or above by 100 adult participants [54], with a Cronbach α internal consistency score of .82 (versus .77 for the full CAFE data set). Subset B contains images showing "substantial variability while minimizing floor and ceiling effects" [54], with a Cronbach α score of .768 (close to the score of .77 for the full data set).

## Results

### Frame Processing

The HollywoodSquares annotators processed 106,001 unique frames (273,493 including the testing phase and 491,343 unique labels when counting multiple labels for the same frame as a different label). Of the 106,001 unique frames labeled, 39,968 received an emotion label corresponding to 1 of the 7 CAFE emotions (not including the testing phase labels). Table 1 contains the number of frames that were included in the training set for each emotion class, including how many children and videos are represented for each emotion category. The frames that were not included received labels of "None" (corresponding to a situation where no face or an incomplete face appears in the frame), "Unknown" (corresponding to the face not expressing a clear emotion), or "Contempt" (corresponding to the face not expressing an emotion in the CAFE set). The large number of curated frames displaying emotion demonstrates the usefulness of HollywoodSquares in filtering out emotion events from noisy data streams. The lack of balance across emotion categories is a testament particularly to the difficulty of evoking anger and sadness as well as disgust and fear, although to a lesser extent.

Of the children who completed 1 session of the Emoji challenge in GuessWhat and uploaded a video to share with the research team, 75 were female, 141 were male, and 51 did not specify their gender. Table 2 presents the racial and ethnic makeup of the participant cohort. Representative GuessWhat frames and cropped faces used to train the classifier, obtained from the subset of participants who consented explicitly to public sharing of their images, are displayed in Figure 3.

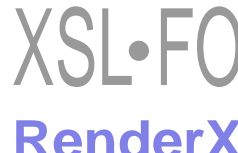

**Table 1.** Emotions represented in the HollywoodSquares data set, including how many children and videos are represented for each emotion category.

| Emotion | Frequency | Number of children | Number of videos |
|---|---|---|---|
| Anger | 643 | 28 | 62 |
| Disgust | 1723 | 46 | 95 |
| Fear | 1875 | 41 | 89 |
| Happy | 13,332 | 73 | 228 |
| Neutral | 16,055 | 87 | 289 |
| Sad | 947 | 31 | 93 |
| Surprise | 5393 | 52 | 135 |

**Table 2.** Representation of race and ethnicity of children whose who played the "Emoji" charades category and uploaded a video to the cloud.

| Race/ethnicity | Frequency |
|---|---|
| Arab | 6 |
| Black or African | 16 |
| East Asian | 16 |
| Hispanic | 36 |
| Native American | 7 |
| Pacific Islander | 5 |
| South Asian | 14 |
| Southeast Asian | 7 |
| White or Caucasian | 100 |
| Not specified | 60 |

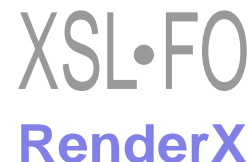

**Figure 3.** Example of frames collected from GuessWhat gameplay, including examples of cropped (A) and original (B) frames. We have displayed these images after obtaining consent from the participants for public sharing.

## Performance on CAFE, CAFE-Defined Subsets, and CAFE Subset Balanced in Terms of Race, Gender, and Emotions

The ResNet-152 network trained on the entire labeled HollywoodSquares data set as well as the JAFFE, AffectNet subset, and CK+ data sets achieved a balanced accuracy of 66.9% and an F1-score of 67.4% on the entirety of the CAFE data set (confusion matrix in Figure 4). When only the HollywoodSquares data set was included in the training set, the model achieved a balanced accuracy of 64.12% and an F1-score of 64.2%. When only including the JAFFE, AffectNet subset, and CK+ sets, the classifier achieved an F1-score of 56.14% and a balanced accuracy of 52.5%, highlighting the contribution of the HollywoodSquares data set.

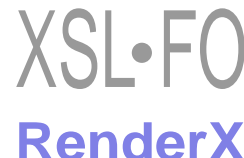

**Figure 4.** Confusion matrix for the entirety of the Child Affective Facial Expression data set.

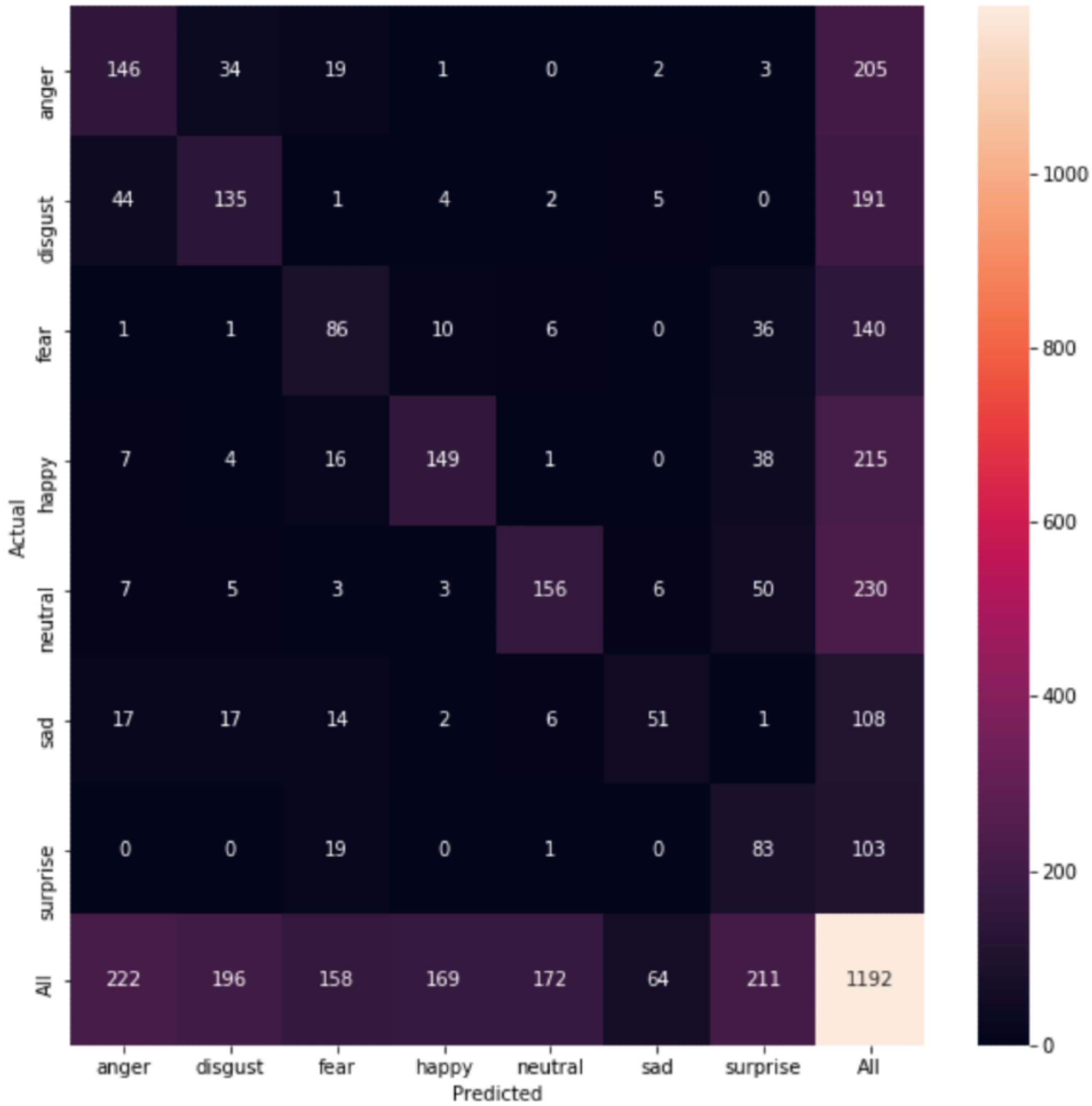

To quantify the contribution of the neural network architecture itself, we compared the performance of several state-of-the-art neural network architectures when only including the HollywoodSquares data set in the training set (Table 3). We evaluated the following models: ResNet152V2 [49], ResNet50V2 [49], InceptionV3 [55], MobileNetV2 [56], DenseNet121 [57], DenseNet201 [57], and Xception [58]. The same training conditions and hyperparameters were used across all models. We found that ResNet152V2 performed better than the other networks when trained with our data, so we used this model for the remainder of our experiments.

The performance improved, resulting in a balanced accuracy of 79.1% and an F1-score of 78% on CAFE Subset A (confusion matrix in Figure 5), a subset containing more universally accepted emotions labels. When only including the non-GuessWhat public images in the training set, the model achieved a balanced accuracy of 65.3% and an F1-score of 69.2%. On CAFE Subset B, the balanced accuracy was 66.4% and the F1-score was 67.2% (confusion matrix in Figure 6); the balanced accuracy was 57.2% and F1-score was 57.3% when exclusively training on the non-GuessWhat public images.

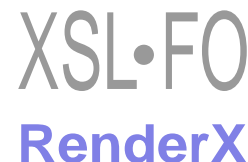

**Table 3.** Comparison of several popular neural network architectures trained on the same data set[a].

| Model | Balanced accuracy (%) | F1-score (%) | Number of network parameters |
|---|---|---|---|
| ResNet152V2; He et al [49] | 64.12 | 64.2 | 60,380,648 |
| ResNet50V2; He et al [49] | 63.67 | 63.12 | 25,613,800 |
| InceptionV3; Szegedy et al [55] | 59 | 59.66 | 23,851,784 |
| MobileNetV2; Sandler et al [56] | 57.63 | 58.19 | 3,538,984 |
| DenseNet121; Huang et al [57] | 58.2 | 59.19 | 8,062,504 |
| DenesNet201; Huang et al [57] | 57.02 | 58.95 | 20,242,984 |
| Xception; Chollet and François [58] | 58.16 | 60.58 | 22,910,480 |

[a]Default hyperparameters were used for all networks.

**Figure 5.** Confusion matrix for Child Affective Facial Expression Subset A.

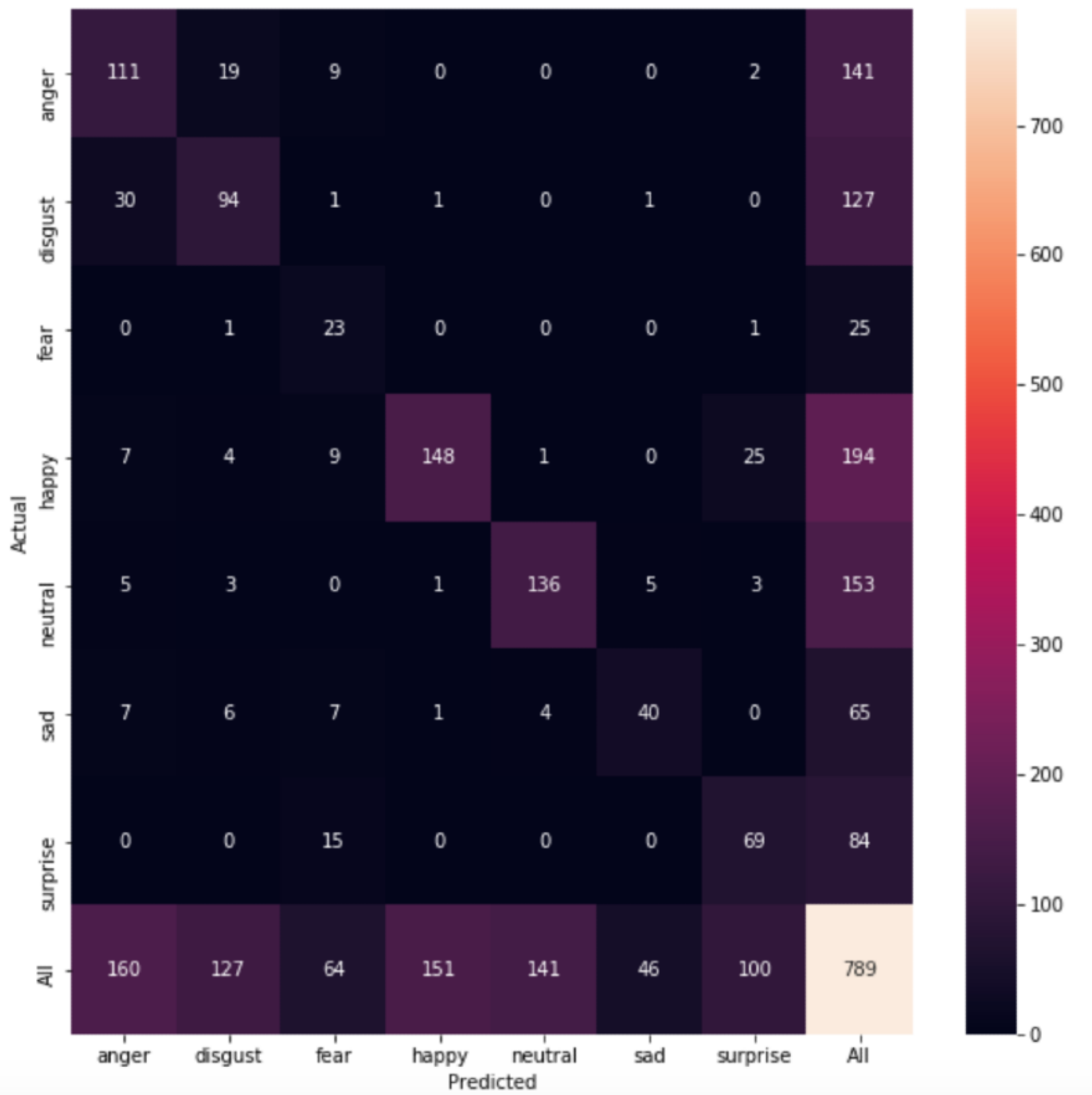

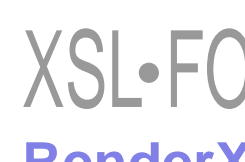

**Figure 6.** Confusion matrix for Child Affective Facial Expression Subset B.

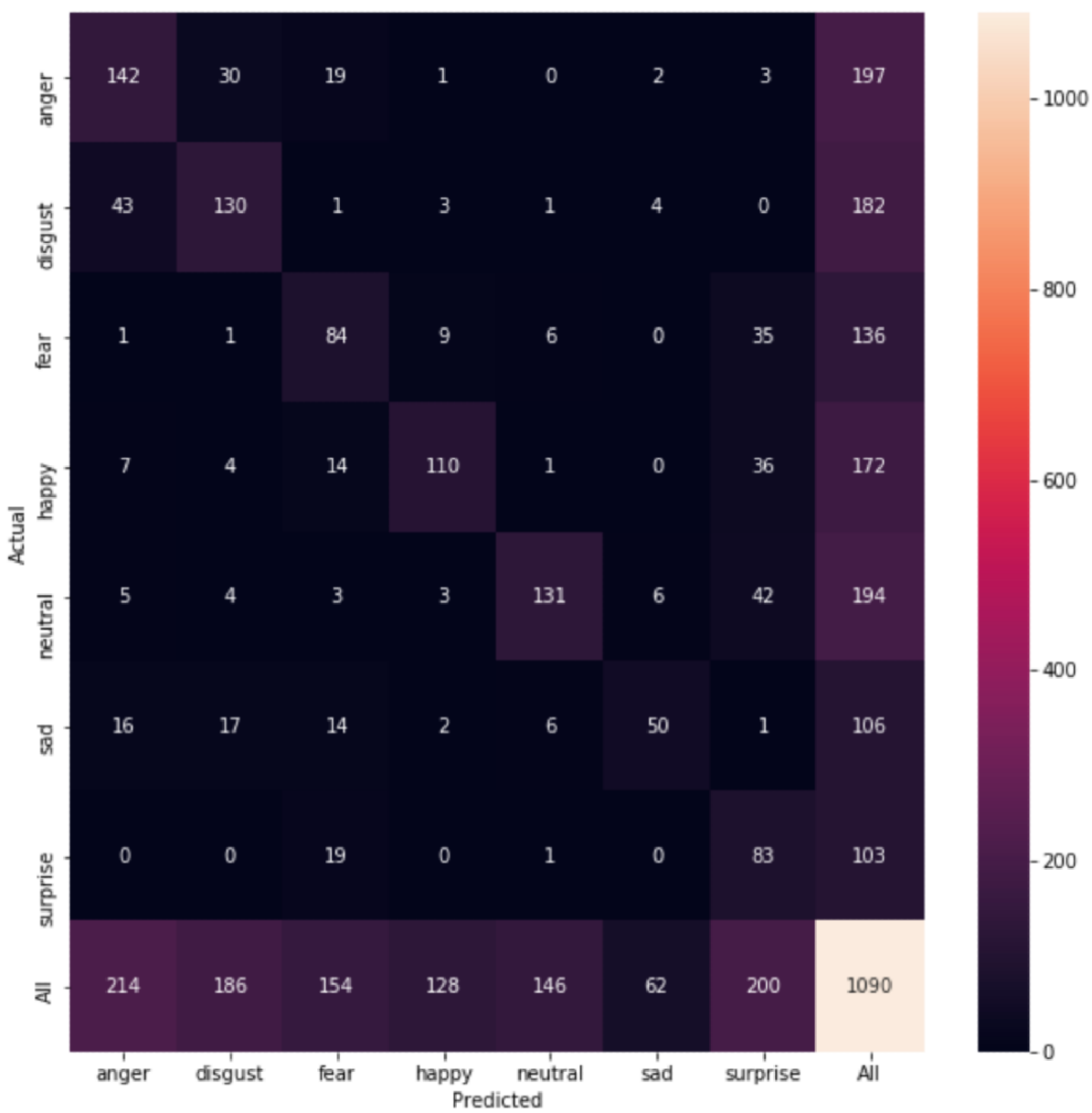

## Classifier Performance Based on Image Difficulty

CAFE images were labeled by 100 adults, and the percentage of participants who labeled the correct class are reported with the data set [54]. We binned frames into 10 difficulty classes (ie, 90%-100% correct human labels, 80%-90% correct human labels, etc). Figure 7 shows that our classifier performs exceedingly well on unambiguous images. Of the 233 images with 90%-100% agreement between the original CAFE labelers, our classifier correctly classifies 90.1% of the images. The true label makeup of these images is as follows: 131 happy, 58 neutral, 20 anger, 9 sad, 8 surprise, 7 disgust, and 0 fear images. This confirms that humans have trouble identifying nonhappy and nonneutral facial expressions. Of the 455 images with 80%-100% agreement between the original CAFE labelers, our classifier correctly classifies 81.1% of the images.

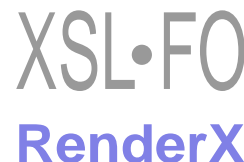

**Figure 7.** Classifier performance versus original CAFE annotator performance for 10 difficulty bins. The classifier tends to perform well when humans agree on the class and poorly otherwise. The numbers in parentheses represent the number of images in each bin. This highlights the issue of ambiguous labels in affective computing and demonstrates that our model performance scales proportionally to human performance. CAFE: Child Affective Facial Expression.

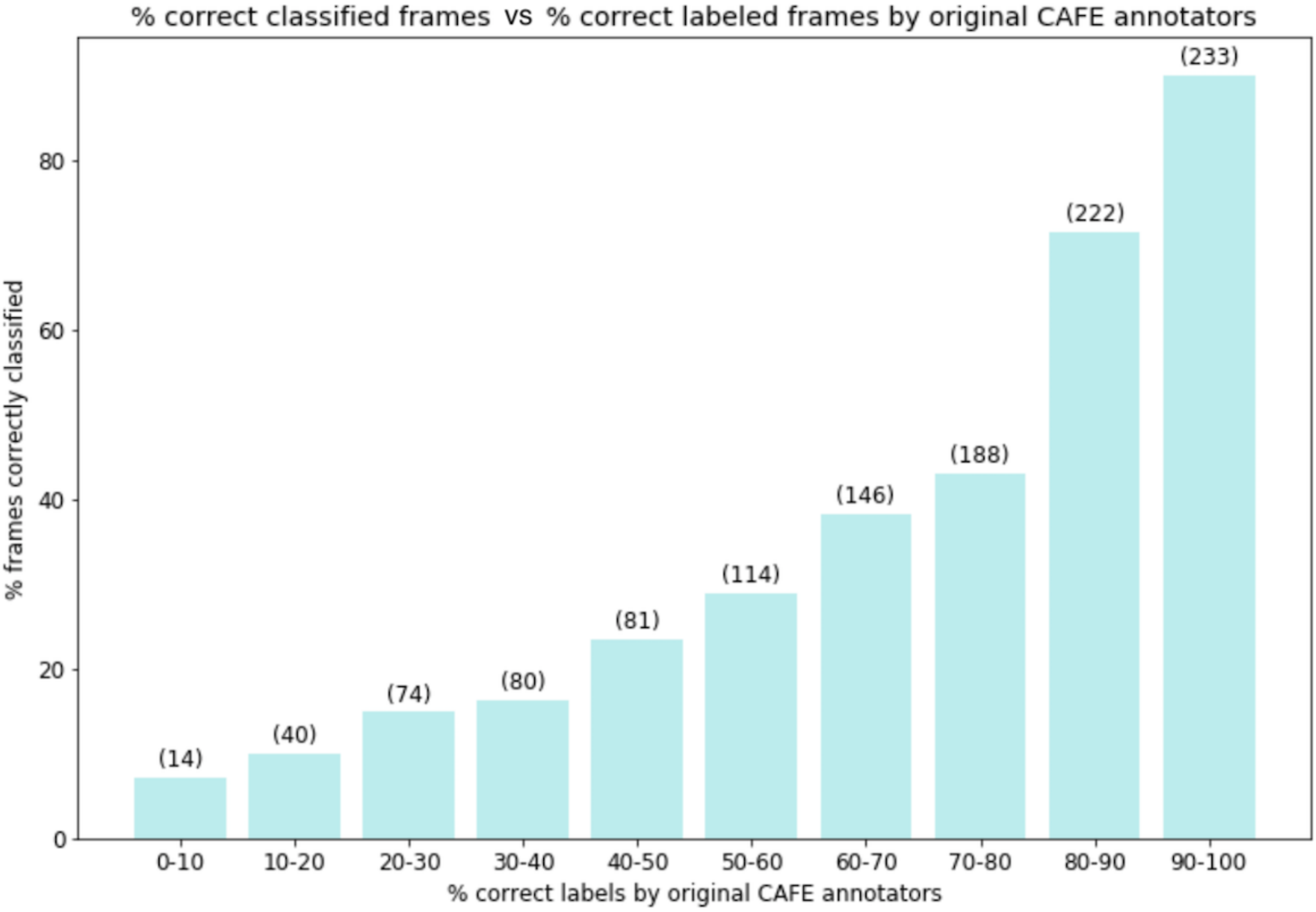

## *Discussion*

### Principal Results

Through the successful application of an in-the-wild child developmental health therapeutic that simultaneously captures video data, we show that a pipeline for intelligently and continuously labeling image frames collected passively from mobile gameplay can generate sufficient training data for a high-performing computer vision classifier (relative to prior work). We curated a data set that contains images enriched for naturalistic facial expressions of children, including but not limited to children with autism.

We demonstrate the best-performing pediatric facial emotion classifier to date according to the CAFE data set. The best-performing classifiers evaluated in earlier studies involving facial emotion classification on the CAFE data set, including images from CAFE in the training set, achieved an accuracy of up to 56% on CAFE [36,37,39] and combined "anger" and "disgust" into a single class. By contrast, we achieved a balanced accuracy of 66.9% and an F1-score of 67.4% without including any CAFE images in the training set. This is a clear illustration of the power of parallel data curation from distributed mobile devices in conjunction with deep learning, and this approach can possibly be generalized to the collection of training data for other domains.

We collected a sufficiently large training sample to alleviate the need for extracting facial keypoint features, as was the case in prior works. Instead, we used the unaltered images as inputs to a deep CNN.

### Limitations and Future Work

A major limitation of this work is the use of 7 discrete and distinct emotion categories. Some images in the training set might have exhibited more than 1 emotion, such as "happily surprised" or "fearfully surprised." This could be addressed in future work by a more thorough investigation of the final emotion classes. Another limitation is that similar to existing emotion data sets, our generated data set contains fake emotion evocations by the children. This is due to limitations imposed by ethics review committees and the IRB who, understandably so, do not allow provoking real fear or sadness in participants, especially young children who may have a developmental delay. This issue of fake emotion evocation has been documented in prior studies [4,5,59,60]. Finding a solution to this issue that would appease ethical review committees is an open research question.

Another limitation is that we did not address the possibility of complex or compound emotions [61]. A particular facial expression can consist of multiple universal expressions. For example, "happily surprised," "fearfully surprised," and even "angrily surprised" are all separate subclasses of "surprised." We have not separated these categories in this study. We recommend that future studies explore the possibility of predicting compound and complex facial expressions.

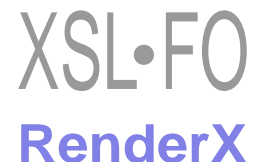

There are several fruitful avenues for future work. The paradigm of passive data collection during mobile intervention gameplay could be expanded to other digital intervention modalities, such as wearable autism systems with front-facing cameras [7,8,11,13-17]. This paradigm can also be applied toward the curation of data and subsequent training of other behavioral classifiers. Relevant computer vision models for diagnosing autism could include computer vision–powered quantification of hand stimming, eye contact, and repetitive behavior, as well as audio-based classification of abnormal prosody, among others.

The next major research step will be to evaluate how systems like GuessWhat can benefit from the incorporation of the machine learning models back into the system in a closed-loop fashion while preserving privacy and trust [62]. Quantification of autistic behaviors during gameplay via machine learning models trained with gameplay videos can enable a feedback loop that provides a dynamic and adaptive therapy for the child. Models can be further personalized to the child's unique characteristics, providing higher performance through customized fine-tuning of the network.

## Conclusions

We have demonstrated that gamified digital therapeutic interventions can generate sufficient data for training state-of-the-art computer vision classifiers, in this case for pediatric facial emotion. Using this data curation and labeling paradigm, we trained a state-of-the-art 7-way pediatric facial emotion classifier.

## Acknowledgments

We would like to acknowledge all the nine high school and undergraduate emotion annotators: Natalie Park, Chris Harjadi, Meagan Tsou, Belle Bankston, Hadley Daniels, Sky Ng-Thow-Hing, Bess Olshen, Courtney McCormick, and Jennifer Yu. The work was supported in part by funds to DPW from the National Institutes of Health (grants 1R01EB025025-01, 1R01LM013364-01, 1R21HD091500-01, and 1R01LM013083), the National Science Foundation (Award 2014232), The Hartwell Foundation, Bill and Melinda Gates Foundation, Coulter Foundation, Lucile Packard Foundation, Auxiliaries Endowment, the Islamic Development Bank Transform Fund, the Weston Havens Foundation, and program grants from Stanford's Human-Centered Artificial Intelligence Program, Precision Health and Integrated Diagnostics Center, Beckman Center, Bio-X Center, Predictives and Diagnostics Accelerator, Spectrum, Spark Program in Translational Research, MediaX, and from the Wu Tsai Neurosciences Institute's Neuroscience:Translate Program. We also acknowledge generous support from David Orr, Imma Calvo, Bobby Dekesyer, and Peter Sullivan. PW would like to acknowledge support from Mr. Schroeder and the Stanford Interdisciplinary Graduate Fellowship (SIGF) as the Schroeder Family Goldman Sachs Graduate Fellow.

## Conflicts of Interest

DPW is the founder of Cognoa.com. This company is developing digital health solutions for pediatric care. AK works as a part-time consultant with Cognoa.com. All other authors declare no conflict of interests.

## References

1. Harms MB, Martin A, Wallace GL. Facial emotion recognition in autism spectrum disorders: a review of behavioral and neuroimaging studies. Neuropsychol Rev 2010 Sep;20:290-322. [doi: 10.1007/s11065-010-9138-6]
2. Hobson RP, Ouston J, Lee A. Emotion recognition in autism: coordinating faces and voices. Psychol Med 2009 Jul;18(4):911-923. [doi: 10.1017/S0033291700009843]
3. Rieffe C, Oosterveld P, Terwogt MM, Mootz S, van Leeuwen E, Stockmann L. Emotion regulation and internalizing symptoms in children with autism spectrum disorders. Autism 2011 Jul;15(6):655-670. [doi: 10.1177/1362361310366571]
4. Carolis BD, D'Errico F, Paciello M, Palestra G. Cognitive emotions recognition in e-learning: exploring the role of age differences and personality traits. In: Methodologies and Intelligent Systems for Technology Enhanced Learning, 9th International Conference. 2019 Jun Presented at: International Conference in Methodologies and intelligent Systems for Technology Enhanced Learning; June 26-28, 2019; Ávila, Spain p. 97-104. [doi: 10.1007/978-3-030-23990-9_12]
5. De Carolis B, D'Errico F, Rossano V. Socio-affective technologies [SI 1156 T]. Multimed Tools Appl 2020 Oct;79:35779-35783. [doi: 10.1007/s11042-020-10015-3]
6. Franzoni V, Biondi G, Perri D, Gervasi O. Enhancing mouth-based emotion recognition using transfer learning. Sensors 2020 Sep;20(18):5222. [doi: 10.3390/s20185222]
7. Daniels J, Haber N, Voss C, Schwartz J, Tamura S, Fazel A, et al. Feasibility testing of a wearable behavioral aid for social learning in children with autism. Appl Clin Inform 2018 Feb;09(01):129-140. [doi: 10.1055/s-0038-1626727]
8. Daniels J, Schwartz JN, Voss C, Haber N, Fazel A, Kline A, et al. Exploratory study examining the at-home feasibility of a wearable tool for social-affective learning in children with autism. NPJ Digital Med 2018 Aug;1:32. [doi: 10.1038/s41746-018-0035-3]
9. Haber N, Voss C, Fazel A, Winograd T, Wall DP. A practical approach to real-time neutral feature subtraction for facial expression recognition. In: 2016 IEEE Winter Conference on Applications of Computer Vision (WACV). 2016 Presented at: IEEE Winter Conference on Applications of Computer Vision (WACV); March 7-10, 2016; Lake Placid, United States p. 1-9. [doi: 10.1109/WACV.2016.7477675]

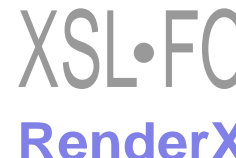

10. Haber N, Voss C, Wall D. Making emotions transparent: Google Glass helps autistic kids understand facial expressions through augmented-reaiity therapy. IEEE Spectrum 2020 Apr;57(4):46-52. [doi: 10.1109/MSPEC.2020.9055973]
11. Kline A, Voss C, Washington P, Haber N, Schwartz H, Tariq Q, et al. Superpower glass. GetMobile: Mobile Comp Comm 2019 Nov;23(2):35-38. [doi: 10.1145/3372300.3372308]
12. Nag A, Haber N, Voss C, Tamura S, Daniels J, Ma J, et al. Toward continuous social phenotyping: analyzing gaze patterns in an emotion recognition task for children with autism through wearable smart glasses. J Med Internet Res 2020 Apr;22(4):e13810. [doi: 10.2196/13810]
13. Voss C, Haber N, Wall DP. The potential for machine learning–based wearables to improve socialization in teenagers and adults with autism spectrum disorder—reply. JAMA Pediatr 2019 Nov;173(11):1106. [doi: 10.1001/jamapediatrics.2019.2969]
14. Voss C, Schwartz J, Daniels J, Kline A, Haber N, Washington P, et al. Effect of wearable digital intervention for improving socialization in children with autism spectrum disorder. JAMA Pediatr 2019 May;173(5):446-454. [doi: 10.1001/jamapediatrics.2019.0285]
15. Voss C, Washington P, Haber N, Kline A, Daniels J, Fazel A, et al. Superpower glass: delivering unobtrusive real-time social cues in wearable systems. In: UbiComp '16: Proceedings of the 2016 ACM International Joint Conference on Pervasive and Ubiquitous Computing: Adjunct. 2016 Sep Presented at: UbiComp '16: The 2016 ACM International Joint Conference on Pervasive and Ubiquitous Computing; September 12-16, 2016; Heidelberg, Germany p. 1218-1226. [doi: 10.1145/2968219.2968310]
16. Washington P, Voss C, Haber N, Tanaka S, Daniels J, Feinstein C, et al. A wearable social interaction aid for children with autism. In: CHI EA '16: Proceedings of the 2016 CHI Conference Extended Abstracts on Human Factors in Computing Systems. 2016 May Presented at: CHI'16: CHI Conference on Human Factors in Computing Systems; May 7-12, 2016; San Jose, United States p. 2348-2354. [doi: 10.1145/2851581.2892282]
17. Washington P, Voss C, Kline A, Haber N, Daniels J, Fazel A, et al. SuperpowerGlass: a wearable aid for the at-home therapy of children with autism. Proc ACM Interact Mob Wearable Ubiquitous Technol 2017 Sep;1(3):1-22. [doi: 10.1145/3130977]
18. Abbas H, Garberson F, Glover E, Wall DP. Machine learning for early detection of autism (and other conditions) using a parental questionnaire and home video screening. 2017 Presented at: IEEE International Conference on Big Data (Big Data); December 11-14, 2017; Boston, United States p. 3558-3561. [doi: 10.1109/bigdata.2017.8258346]
19. Abbas H, Garberson F, Liu-Mayo S, Glover E, Wall DP. Multi-modular AI approach to streamline autism diagnosis in young children. Sci Rep 2020 Mar;10:5014. [doi: 10.1038/s41598-020-61213-w]
20. Duda M, Kosmicki JA, Wall DP. Testing the accuracy of an observation-based classifier for rapid detection of autism risk. Transl Psychiatry 2014 Aug;4:e424. [doi: 10.1038/tp.2014.65]
21. Duda M, Ma R, Haber N, Wall DP. Use of machine learning for behavioral distinction of autism and ADHD. Transl Psychiatry 2016 Feb;6:e732. [doi: 10.1038/tp.2015.221]
22. Duda M, Haber N, Daniels J, Wall DP. Crowdsourced validation of a machine-learning classification system for autism and ADHD. Transl Psychiatry 2017 May;7:e1133. [doi: 10.1038/tp.2017.86]
23. Fusaro VA, Daniels J, Duda M, DeLuca TF, D'Angelo O, Tamburello J, et al. The potential of accelerating early detection of autism through content analysis of YouTube videos. PLoS ONE 2014 Apr;9(4):e93533. [doi: 10.1371/journal.pone.0093533]
24. Levy S, Duda M, Haber N, Wall DP. Sparsifying machine learning models identify stable subsets of predictive features for behavioral detection of autism. Mol Autism 2017 Dec;8:65. [doi: 10.1186/s13229-017-0180-6]
25. Leblanc E, Washington P, Varma M, Dunlap K, Penev Y, Kline A, et al. Feature replacement methods enable reliable home video analysis for machine learning detection of autism. Sci Rep 2020 Dec;10:21245. [doi: 10.1038/s41598-020-76874-w]
26. Stark DE, Kumar RB, Longhurst CA, Wall DP. The quantified brain: a framework for mobile device-based assessment of behavior and neurological function. Appl Clin Inform 2017 Dec;07(02):290-298. [doi: 10.4338/ACI-2015-12-LE-0176]
27. Tariq Q, Daniels J, Schwartz JN, Washington P, Kalantarian H, Wall DP. Mobile detection of autism through machine learning on home video: a development and prospective validation study. PLoS Med 2018 Nov;15(11):e1002705. [doi: 10.1371/journal.pmed.1002705]
28. Tariq Q, Fleming SL, Schwartz JN, Dunlap K, Corbin C, Washington P, et al. Detecting developmental delay and autism through machine learning models using home videos of Bangladeshi children: development and validation study. J Med Internet Res 2019 Apr;21(4):e13822. [doi: 10.2196/13822]
29. Washington P, Kalantarian H, Tariq Q, Schwartz J, Dunlap K, Chrisman B, et al. Validity of online screening for autism: crowdsourcing study comparing paid and unpaid diagnostic tasks. J Med Internet Res 2019 May;21(5):e13668. [doi: 10.2196/13668]
30. Washington P, Leblanc E, Dunlap K, Penev Y, Kline A, Paskov K, et al. Precision telemedicine through crowdsourced machine learning: testing variability of crowd workers for video-based autism feature recognition. J Pers Med 2020 Aug;10(3):86. [doi: 10.3390/jpm10030086]
31. Washington P, Leblanc E, Dunlap K, Penev Y, Varma M, Jung JY, et al. Selection of trustworthy crowd workers for telemedical diagnosis of pediatric autism spectrum disorder. Biocomputing 2021: Proceedings of the Pacific Symposium 2020:14-25. [doi: 10.1142/9789811232701_0002]

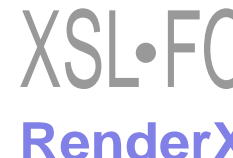

32. Washington P, Paskov KM, Kalantarian H, Stockham N, Voss C, Kline A, et al. Feature selection and dimension reduction of social autism data. Biocomputing 2020 2020:707-718. [doi: 10.1142/9789811215636_0062]
33. Washington P, Park N, Srivastava P, Voss C, Kline A, Varma M, et al. Data-driven diagnostics and the potential of mobile artificial intelligence for digital therapeutic phenotyping in computational psychiatry. Biol Psychiatry Cogn Neurosci Neuroimaging 2020 Aug;5(8):759-769. [doi: 10.1016/j.bpsc.2019.11.015]
34. Baker L, LoBue V, Bonawitz E, Shafto P. Towards automated classification of emotional facial expressions. CogSci 2017:1574-1579.
35. Florea C, Florea L, Badea M, Vertan C, Racoviteanu A. Annealed label transfer for face expression recognition. BMVC 2019:104. [doi: 10.1109/ecai50035.2020.9223242]
36. Lopez-Rincon A. Emotion recognition using facial expressions in children using the NAO Robot. 2019 Presented at: 2019 International Conference on Electronics, Communications and Computers (CONIELECOMP); February 27- March 1, 2019; Cholula, Mexico p. 146-153. [doi: 10.1109/CONIELECOMP.2019.8673111]
37. Nagpal S, Singh M, Vatsa M, Singh R, Noore A. Expression classification in children using mean supervised deep Boltzmann machine. 2019 Jun Presented at: 2019 IEEE/CVF Conference on Computer Vision and Pattern Recognition Workshops (CVPRW); June 16-17, 2019; Long Beach, United States p. 236-245. [doi: 10.1109/CVPRW.2019.00033]
38. Rao A, Ajri S, Guragol A, Suresh R, Tripathi S. Emotion recognition from facial expressions in children and adults using deep neural network. Intelligent Systems, Technologies and Applications 2020 May:43-51. [doi: 10.1007/978-981-15-3914-5_4]
39. Witherow MA, Samad MD, Iftekharuddin KM. Transfer learning approach to multiclass classification of child facial expressions. Applications of Machine Learning 2019 Sep:1113911. [doi: 10.1117/12.2530397]
40. Kalantarian H, Jedoui K, Washington P, Tariq Q, Dunlap K, Schwartz J, et al. Labeling images with facial emotion and the potential for pediatric healthcare. Artif Intell Med 2019 Jul;98:77-86. [doi: 10.1016/j.artmed.2019.06.004]
41. Kalantarian H, Jedoui K, Washington P, Wall DP. A mobile game for automatic emotion-labeling of images. IEEE Trans Games 2020 Jun;12(2):213-218. [doi: 10.1109/TG.2018.2877325]
42. Kalantarian H, Washington P, Schwartz J, Daniels J, Haber N, Wall D. A gamified mobile system for crowdsourcing video for autism research. 2018 Jul Presented at: 2018 IEEE International Conference on Healthcare Informatics (ICHI); June 4-7, 2018; New York City, United States p. 350-352. [doi: 10.1109/ICHI.2018.00052]
43. Kalantarian H, Washington P, Schwartz J, Daniels J, Haber N, Wall DP. Guess What? J Healthc Inform Res 2018 Oct;3:43-66. [doi: 10.1007/s41666-018-0034-9]
44. Kalantarian H, Jedoui K, Dunlap K, Schwartz J, Washington P, Husic A, et al. The performance of emotion classifiers for children with parent-reported autism: quantitative feasibility study. JMIR Ment Health 2020 Apr;7(4):e13174. [doi: 10.2196/13174]
45. Ekman P. Are there basic emotions? Psychological Rev 1992;99(3):550-553. [doi: 10.1037/0033-295x.99.3.550]
46. Ekman P, Scherer KR, editors. Expression and the nature of emotion. In: Approaches to Emotion. United Kingdom: Taylor & Francis; 1984.
47. Molnar P, Segerstrale U, editors. Universal facial expressions of emotion. In: Nonverbal Communication: Where Nature Meets Culture. United Kingdom: Routledge; 1997:27-46.
48. Ekman P, Friesen WV. Constants across cultures in the face and emotion. J Pers Soc Psychol 1971;17(2):124-129. [doi: 10.1037/h0030377]
49. He K, Zhang X, Ren S, Sun J. Deep residual learning for image recognition. 2016 Dec Presented at: 2016 IEEE Conference on Computer Vision and Pattern Recognition (CVPR); June 27-30, 2016; Las Vegas, United States p. 770-778. [doi: 10.1109/CVPR.2016.90]
50. Deng J, Dong W, Socher R, Li LJ, Li K, Fei-Fei L. Imagenet: a large-scale hierarchical image database. In: 2009 IEEE Conference on Computer Vision and Pattern Recognition. 2009 Presented at: IEEE Conference on Computer Vision and Pattern Recognition; June 20-25, 2009; Miami, United States p. 248-255. [doi: 10.1109/CVPR.2009.5206848]
51. Lyons M, Akamatsu S, Kamachi M, Gyoba J. Coding facial expressions with gabor wavelets. 1998 Presented at: Proceedings Third IEEE International Conference on Automatic Face and Gesture Recognition; April 14-16, 1998; Nara, Japan p. 200-205. [doi: 10.1109/afgr.1998.670949]
52. Mollahosseini A, Hasani B, Mahoor MH. AffectNet: a database for facial expression, valence, and arousal computing in the wild. IEEE Trans Affective Comput 2019 Jan;10(1):18-31. [doi: 10.1109/TAFFC.2017.2740923]
53. Lucey P, Cohn JF, Kanade T, Saragih J, Ambadar Z, Matthews I. The extended cohn-kanade dataset (ck+): a complete dataset for action unit and emotion-specified expression. 2010 Aug Presented at: 2010 IEEE Computer Society Conference on Computer Vision and Pattern Recognition - Workshops; June 13-18, 2010; San Francisco, United States p. 94-101. [doi: 10.1109/CVPRW.2010.5543262]
54. LoBue V, Thrasher C. The Child Affective Facial Expression (CAFE) set: validity and reliability from untrained adults. Front Psychol 2015 Jan;5:1532. [doi: 10.3389/fpsyg.2014.01532]
55. Szegedy C, Vanhoucke V, Ioffe S, Shlens J, Wojna Z. Rethinking the inception architecture for computer vision. 2016 Presented at: 2016 IEEE Conference on Computer Vision and Pattern Recognition (CVPR); June 27-30, 2016; Las Vegas, United States p. 2818-2826. [doi: 10.1109/cvpr.2016.308]

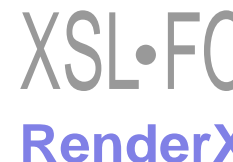

56. Sandler M, Howard A, Zhu M, Zhmoginov A, Chen LC. Mobilenetv2: inverted residuals and linear bottlenecks. 2018 Presented at: 2018 IEEE/CVF Conference on Computer Vision and Pattern Recognition; June 18-23, 2018; Salt Lake City, United States p. 4510-4520. [doi: 10.1109/CVPR.2018.00474]
57. Huang G, Liu Z, Van Der Maaten L, Weinberger KQ. Densely connected convolutional networks. 2017 Nov Presented at: 2017 IEEE Conference on Computer Vision and Pattern Recognition (CVPR); July 21-26, 2017; Honolulu, United States p. 2261-2269. [doi: 10.1109/CVPR.2017.243]
58. Chollet F. Xception: deep learning with depthwise separable convolutions. In: Proceedings of the IEEE Conference on Computer Vision and Pattern Recognition. 2017 Nov Presented at: IEEE Conference on Computer Vision and Pattern Recognition; July 21-26, 2017; Honolulu, United States p. 1251-1258. [doi: 10.1109/CVPR.2017.195]
59. Dawel A, Wright L, Irons J, Dumbleton R, Palermo R, O'Kearney R, et al. Perceived emotion genuineness: normative ratings for popular facial expression stimuli and the development of perceived-as-genuine and perceived-as-fake sets. Behav Res 2016 Dec;49:1539-1562. [doi: 10.3758/s13428-016-0813-2]
60. Vallverdú J, Nishida T, Ohmoto Y, Moran S, Lázare S. Fake empathy and human-robot interaction (HRI): A preliminary study. IJTHI 2018;14(1):44-59. [doi: 10.4018/IJTHI.2018010103]
61. Du S, Tao Y, Martinez AM. Compound facial expressions of emotion. Proc Natl Acad Sci 2014 Mar;111(15):E1454-E1462. [doi: 10.1073/pnas.1322355111]
62. Washington P, Yeung S, Percha B, Tatonetti N, Liphardt J, Wall DP. Achieving trustworthy biomedical data solutions. Biocomputing 2021: Proceedings of the Pacific Symposium 2021:1-13. [doi: 10.1142/9789811232701_0001]

## Abbreviations

**AWS:** Amazon Web Services
**CAFE:** Child Affective Facial Expression data set
**CK+:** Extended Cohn-Kanade data set
**CNN:** convolutional neural network
**GPU:** graphics processing unit
**IRB:** Institutional Review Board
**JAFFE:** Japanese Female Facial Expression data set

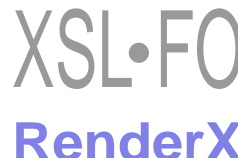